\documentclass{article}
\usepackage{amsmath,epsfig}
\usepackage[preprint]{spconfa4}
\usepackage{cite}

\usepackage{times}
\usepackage{graphicx}
\usepackage{amssymb}
\usepackage{multirow}
\usepackage{subcaption}
\usepackage{array}

\usepackage[export]{adjustbox}

\usepackage{booktabs}
\usepackage{comment}
\usepackage{algorithm,algpseudocode}
\usepackage{url}
\usepackage{bbding}
\usepackage{verbatim}
\usepackage{wasysym} 
\usepackage[table,x11names]{xcolor}
\definecolor{mygray}{gray}{0.95}
\usepackage{amsfonts}
\usepackage{bm}
\usepackage{mathtools}
\usepackage{color}
\let\OLDthebibliography\thebibliography
\renewcommand\thebibliography[1]{
  \OLDthebibliography{#1}
  \setlength{\parskip}{0pt}
  \setlength{\itemsep}{0pt plus 0.3ex}
}

\begin{document}\sloppy

\def\x{{\mathbf x}}
\def\L{{\cal L}}

\title{Causal Video Summarizer for Video Exploration}
%
\name{Jia-Hong Huang$^{1}$, Chao-Han Huck Yang$^{2}$, Pin-Yu Chen$^{3}$, Andrew Brown$^{1}$, Marcel Worring$^{1}$}
\address{$^{1}$University of Amsterdam, Netherlands $^{2}$Georgia Institute of Technology, USA $^{3}$IBM Research, USA \\
{\tt\small j.huang@uva.nl, huckiyang@gatech.edu, pin-yu.chen@ibm.com, a.g.brown@uva.nl, m.worring@uva.nl}}


\maketitle

\begin{abstract}
Recently, video summarization has been proposed as a method to help video exploration. However, traditional video summarization models only generate a fixed video summary which is usually independent of user-specific needs and hence limits the effectiveness of video exploration. Multi-modal video summarization is one of the approaches utilized to address this issue. Multi-modal video summarization has a video input and a text-based query input. Hence, effective modeling of the interaction between a video input and text-based query is essential to multi-modal video summarization. In this work, a new causality-based method named Causal Video Summarizer (CVS) is proposed to effectively capture the interactive information between the video and query to tackle the task of multi-modal video summarization. The proposed method consists of a probabilistic encoder and a probabilistic decoder. Based on the evaluation of the existing multi-modal video summarization dataset, experimental results show that the proposed approach is effective with the increase of $+5.4$\% in accuracy and $+4.92$\% increase of $F1$-score, compared with the state-of-the-art method.
\end{abstract}

\section{Introduction}
Video contents are growing at an ever-increasing speed and beyond the capacity of an individual for full comprehension. According to \cite{ji2019video}, more than 18,000 hours of videos are uploaded to YouTube per minute. Exploring this quantity of video data is a daunting task, and video summarization is gaining traction as the ideal solution \cite{potapov2014category,song2015tvsum}.
The main idea of video summarization is to automatically generate a short video clip that summarizes the content of an original, longer video by capturing its important parts \cite{gygli2015video,zhang2016summary,zhang2019dtr}. However, traditional video summarization approaches, e.g., \cite{de2011vsumm,sharghi2018improving}, only create a fixed video summary for a given video.
Given that viewers may have different preferences and videos have different ideal summarization, this issue reduces the effectiveness of video exploration. 


\begin{figure}[t]
\begin{center}
\includegraphics[width=1.0\linewidth]{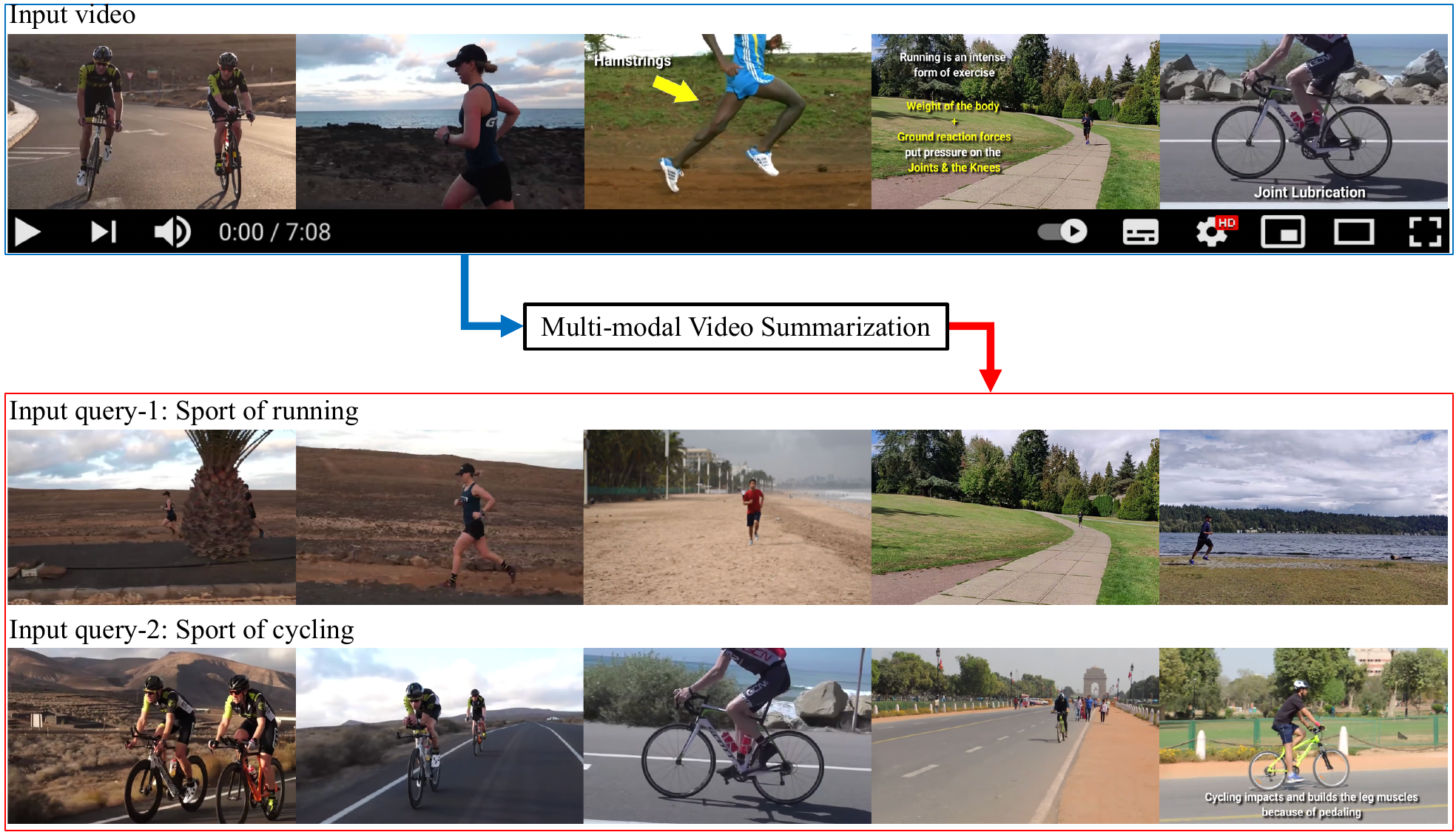}
\end{center}
\vspace{-0.6cm}
   \caption{Multi-modal video summarization. The input video is summarised taking into account text-based queries. ``Input query-1: Sport of running'' and ``Input query-2: Sport of cycling' independently drive the model to generate video summaries that contain running-related content and cycling-related content, respectively.}
\vspace{-0.5cm}
\label{fig:figure11}
\end{figure}

Multi-modal video summarization has been proposed as an approach to improve the effectiveness of video exploration \cite{vasudevan2017query,huang2020query}. It generates video summaries for a given video based on the text-based query provided by the user, illustrated in Figure \ref{fig:figure11}. Conventional video summarization only has video input modality, while an efficient choice for multi-modal video summarization is a text-based query, in addition to video \cite{vasudevan2017query,huang2020query}. 
When multi-modal video summarization is used to help video exploration, effectively modeling the implicit relation/interaction between the text-based query and the video is important \cite{vasudevan2017query}. In \cite{huang2020query}, the proposed model exploits a joint representation of vision and language to perform multi-modal video summarization. However, the implicit interaction between the query and the video is not properly modeled because the simple average-based method used in \cite{huang2020query} probably is not effective enough.

As stated in \cite{louizos2017causal,yang2022training,yang2022treatment}, causal effect modeling, visualized in Figure \ref{fig:figure1}, is helpful for a machine learning task, e.g., image/frame classification, and affects model performance in a positive way. In this work, a new causal video summarizer (CVS) is proposed that tackles the aforementioned issue to improve the performance of a multi-modal video summarization model. The proposed causality-based model consists of a multi-modal feature processing module (MFPM), probabilistic encoding module (PEM), and a probabilistic decoding module (PDM), referring to Figure \ref{fig:figure2}.
Studying multi-modal video summarization from the causality perspective \cite{louizos2017causal} eliminates the need for a priori definition of objectives \cite{vasudevan2017query} based on high-level concepts. For the implicit interaction between the video and query, an attention mechanism is applied to better capture the interactive information.

According to \cite{apostolidis2021video}, typically the generated video summary by a video summarization algorithm is composed of a set of representative video frames or video fragments. Frame-based video summaries are not restricted by timing or synchronization issues and, therefore, they provide more flexibility in terms of data organization for video exploration purpose \cite{calic2007efficient,apostolidis2021video}. In this work, the proposed CVS is validated on the frame-based multi-modal video summarization dataset \cite{huang2020query}. Experimental results show that the proposed method is effective and significantly increases both the accuracy and $F1$-score, compared with the state-of-the-art method.




\section{Related Work}
\noindent\textbf{2.1 Video Summarization with A Single Modality}

Recently, a number of methods with different techniques have been proposed for video summarization, such as \cite{gygli2015video,plummer2017enhancing,zhang2019dilated,ji2019video,song2015tvsum}. 
In \cite{gygli2015video,plummer2017enhancing}, human-explainable concepts, e.g., interestingness and representativeness, are used to characterize a good video summary. Then, objectives are built based on those concepts. Visual and textual information related to a video can then be captured by the targeted objectives. However, a notable limitation is that it is hard to define/measure all possible confounders \cite{louizos2017causal} a priori, such as compactness \cite{zhang2019dilated}, uniformity, and representativeness \cite{gygli2015video}. In \cite{ji2019video,zhang2019dilated}, an attention mechanism, a generative adversarial network, and reinforcement learning are used to perform video summarization. 
Since video summarization with a single modality cannot effectively help video exploration \cite{vasudevan2017query,huang2020query}, in this work a new multi-modal video summarization model is proposed to address this problem.

\begin{figure}[t]
\begin{center}
\includegraphics[width=1.0\linewidth]{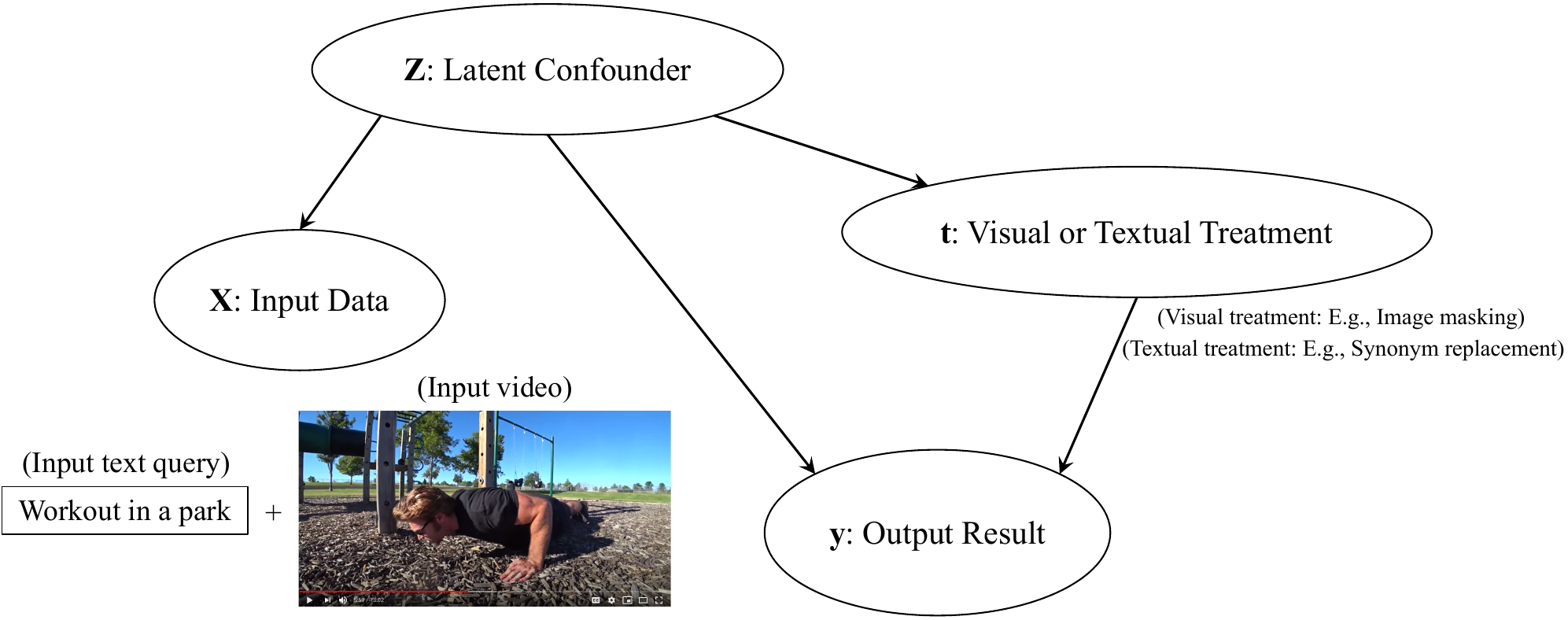}
\end{center}
\vspace{-0.60cm}
  \caption{Explanation of the causal effect modeling concept, i.e., causal graphical model \cite{louizos2017causal}, in multi-modal video summarization. $\textbf{t}$ is a treatment, e.g., visual or textual perturbation \cite{yang2022training}. $\textbf{y}$ is an outcome, e.g., an importance score of a video frame or a relevance score between the input text-based query and video. $\textbf{Z}$ is an unobserved confounder, e.g., representativeness or interestingness \cite{gygli2015video,plummer2017enhancing,vasudevan2017query}. $\textbf{X}$ is a noisy view~\cite{louizos2017causal} on the hidden confounder $\textbf{Z}$, say the input text query and video.}
\vspace{-0.5cm}
\label{fig:figure1}
\end{figure}


\begin{figure*}[t!]
\begin{subfigure}[b]{0.72\textwidth}
\includegraphics[width=1.0\textwidth]{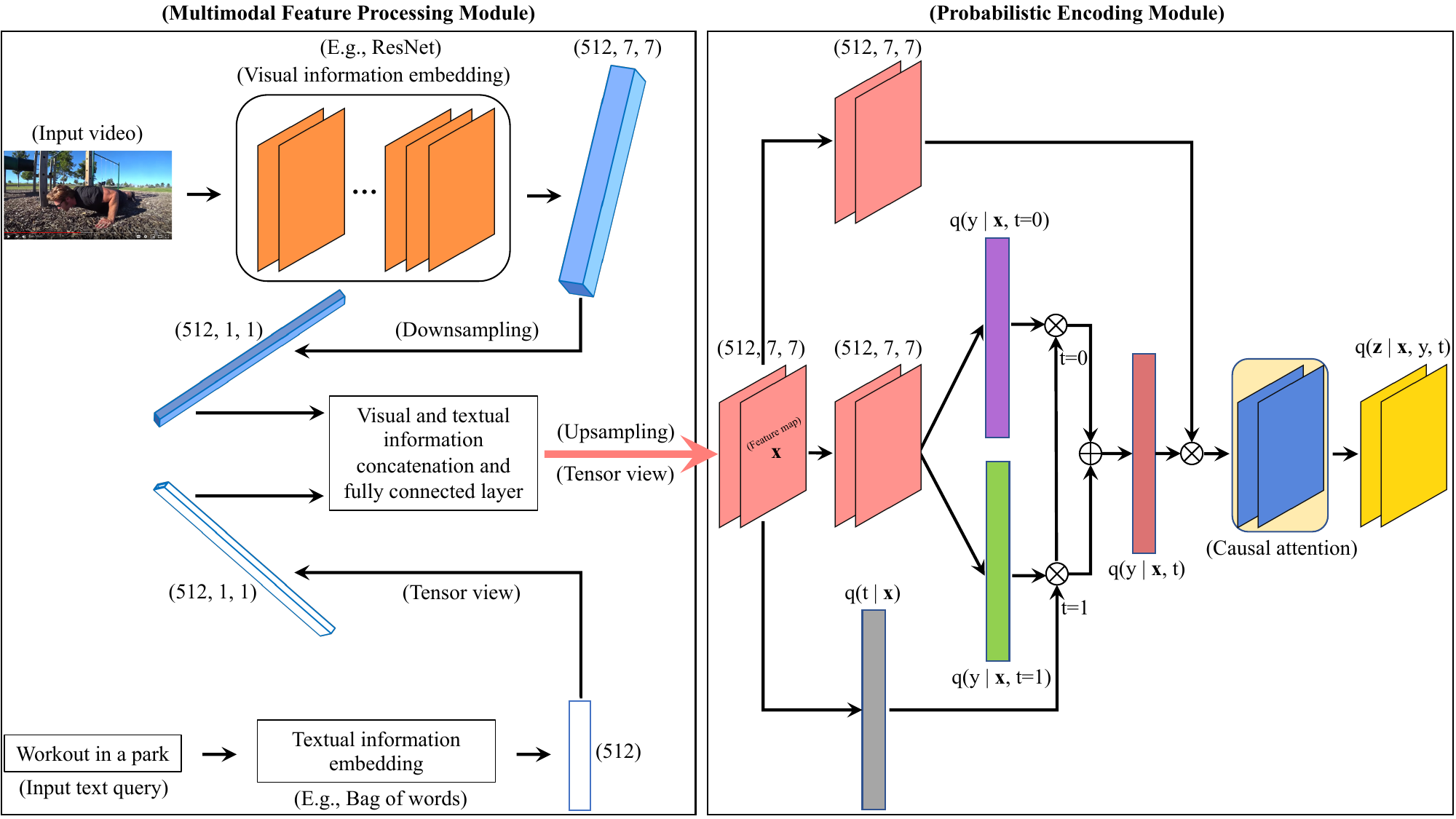}
\caption{Acting as an encoder.}
\end{subfigure}
\begin{subfigure}[b]{0.2585\textwidth}
\includegraphics[width=1.0\textwidth]{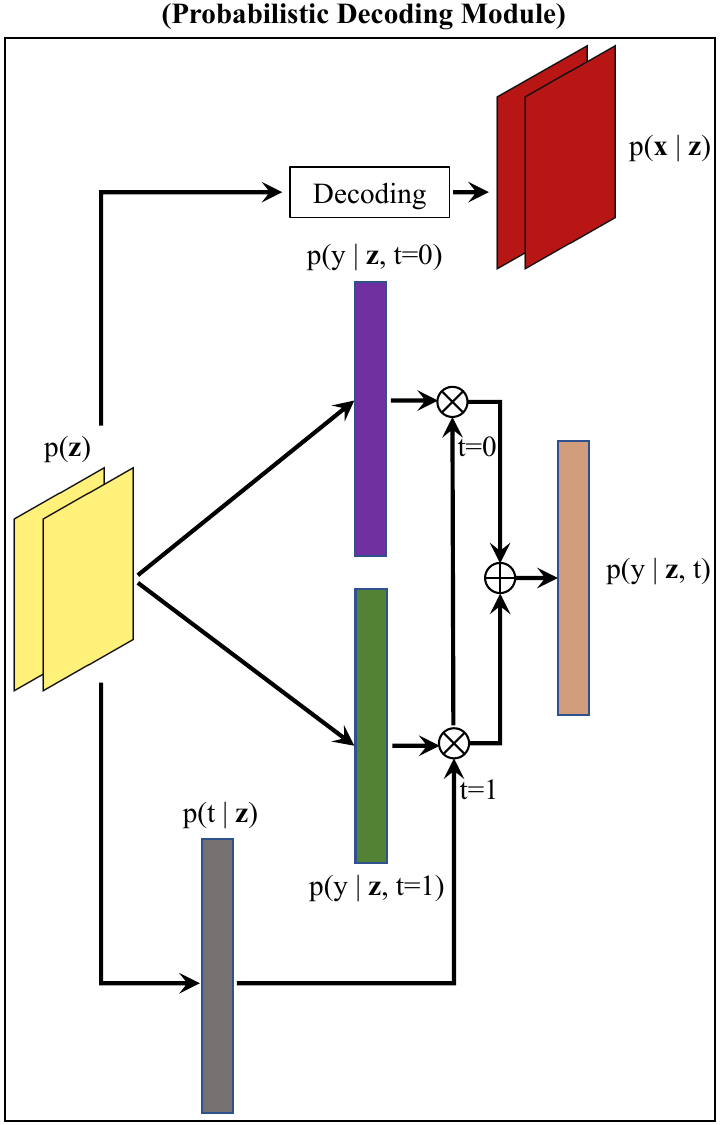}
\caption{Acting as a decoder.}
\end{subfigure}
\vspace{-0.3cm}
   \caption{This figure shows the proposed end-to-end causal attention model for multi-modal video summarization, the practical implementation of the causal graphical model in Figure \ref{fig:figure1}. 
   In (a), the input video and input text query are processed by MFPM for generating the feature map $\textbf{x}$ which is the input of PEM. 
   Then, PEM generates the corresponding probabilistic encoding $q(\textbf{z}|\textbf{x}, y, t)$ of $\textbf{x}$ with the operations of the introduced causal attention. 
   In (b), PDM takes $p(\textbf{z})$, estimated by $q(\textbf{z}|\textbf{x}, y, t)$, to approximately reconstruct $\textbf{x}$ based on $p(\textbf{x}|\textbf{z})$ and generate the prediction score labels $y$ based on $p(y|\textbf{z}, t)$ at the same time. 
   Please refer to \textit{Methodology} section for details.
   }
\vspace{-0.5cm}
\label{fig:figure2}
\end{figure*}

\noindent\textbf{2.2 Multi-modal Video Summarization}

Instead of only considering the visual input, several works have investigated the potential of using additional modalities, e.g., viewers’ comments, video captions, or any other available contextual data, to help models' performance. \cite{sanabria2019deep,wei2018video,huang2020query,huang2021gpt2mvs,huang2019assessing,vasudevan2017query,huang2019novel,huang2017vqabq,hu2019silco,li2017extracting,huang2017robustness,huang2021contextualized,huang2021deep,huang2021longer,huck2018auto,liu2018synthesizing,yang2018novel,di2021dawn,huang2017robustnessMS,huang2021deepopht,huang2022non}. In \cite{li2017extracting}, a multi-modal video summarization method is introduced for key-frame extraction from first-person videos. In \cite{sanabria2019deep}, a multi-modal deep-learning-based model is proposed to summarize videos of soccer games. 
In \cite{wei2018video}, a semantic-based video fragment selection and a visual-to-text mapping are applied based on the relevance between the original and the automatically-generated video descriptions, with the help of semantic attended networks. 
Existing multi-modal video summarization models do not focus on causal effect modeling \cite{louizos2017causal}. Hence, the interactive information between the video and query is difficult to capture effectively. In this work, a new causality-based multi-modal video summarization method is introduced, based on a probabilistic encoder-decoder framework and attention mechanism, to better capture the interaction between the video and query and make the video exploration more effective.


\noindent\textbf{2.3 Causality and Variational Autoencoders}

Proxy variables, e.g., input data $\textbf{X}$ in Figure \ref{fig:figure1}, and the challenges about how to use them correctly have been considered in the literature of causal inference \cite{frost1979proxy}. In many observational studies \cite{gygli2015video,plummer2017enhancing,zhang2020causal}, understanding the best way to derive and measure possible proxy variables is crucial. Building on the previous work \cite{greenland1983correcting,selen1986adjusting}, the authors of \cite{louizos2017causal} and recent works \cite{abbasnejad2020counterfactual, zhang2020causal} for vision applications have tried to exploit proxy variables to study conditions for causal identifiability. In many cases, the general idea is that one should first attempt to infer the joint distribution $p(\textbf{X}, \textbf{Z})$ between the hidden confounders and the proxy, and then use that knowledge to adjust for the hidden confounders \cite{miao2018identifying,pearl2012measurement,kuroki2014measurement}. Take the causal graphical model in Figure \ref{fig:figure1} as an example. The authors of \cite{pearl2012measurement} have shown that if $\textbf{X}$ and $\textbf{Z}$ are categorical, with $\textbf{X}$ having at least as many categories as $\textbf{Z}$, and with the matrix $p(\textbf{X}, \textbf{Z})$ being full-rank, one could use a matrix inversion formula, an approach called ``effect restoration'' \cite{kuroki2014measurement}, to identify the causal effect of $\textbf{t}$ on $\textbf{y}$, referring to \textit{Methodology} section for details. Recently, the authors of \cite{miao2018identifying} have given the conditions under which one could identify more complicated and general proxy models. The proposed causality-based method for multi-modal video summarization is mainly inspired by \cite{louizos2017causal,kingma2013auto,pearl2012measurement,frost1979proxy}.

\section{Methodology}

\noindent\textbf{3.1 Causal Video Summarizer (CVS)}

In this section, 
details of the proposed CVS are described. Note that causal effect modeling for real-world multi-modal video summarization is very complicated \cite{louizos2017causal}. Hence, in this work, the assumptions mentioned in \cite{louizos2017causal} are followed to model the problem of multi-modal video summarization based on the concept of causal effect inference, i.e., causal graphical model. The proposed CVS is mainly composed of a multi-modal feature processing module (MFPM), a probabilistic encoding module (PEM), and a probabilistic decoding module (PDM), referring to Figure \ref{fig:figure2}. The generation of a good video summary is affected by latent factors. In this work, the latent factors are considered as the causal effect. Specifically, the concept of causal effect inference \cite{louizos2017causal}, i.e., a causal graphical model, is used to model the multi-modal video summarization problem, referring to Figure \ref{fig:figure1}. Figure \ref{fig:figure1} contains the four key components of the causal graphical model: ``input data ($\textbf{X}$)'', ``latent confounder ($\textbf{Z}$)'', ``treatment ($\textbf{t}$)'', and ``output result ($\textbf{y}$)''. From the modeling perspective of multi-modal video summarization, $\textbf{X}$ is ``an input video with a text query''. $\textbf{t}$ is ``a visual or textual treatment''. Note that treatment in causal effect modeling is a way to make an input characteristic more salient and help a model learn better \cite{louizos2017causal}. $\textbf{y}$ is ``a relevance score between the input text-based query and video frame or an importance score of a video frame''. $\textbf{Z}$ is ``a variational latent representation'' which the proposed model aims to learn from $\textbf{X}$ for reconstruction \cite{kingma2013auto,louizos2017causal}, referring to Figure \ref{fig:figure2} for the practical implementation. The proposed causal model simultaneously generates outcome score labels $\textbf{y}$ when $\textbf{X}$ is reconstructed, illustrated in Figure \ref{fig:figure2}. Thereafter, video summaries can be created based on the generated outcome score labels.

\noindent\textbf{Objective function for training.}
According to \cite{kingma2013auto,louizos2017causal, shalit2017estimating}, variational autoencoder (VAE) aims to learn a variational latent representation $\textbf{Z}$ from data $\textbf{X}$ for reconstruction; it is capable of learning the latent variables when used in a causal graphical model. Hence, the proposed CVS is built on top of VAE.
Now, it is time to form a single objective for the encoder, in Figure \ref{fig:figure2}-(a), and the decoder, in Figure \ref{fig:figure2}-(b), to learn meaningful causal representations in the latent space and generate video summaries. Based on Figure \ref{fig:figure1}, we know that the true posterior over $\textbf{Z}$ depends on not just $\textbf{X}$, but also on $\textbf{t}$ and $\textbf{y}$. We have to know the treatment assignment $\textbf{t}$ along with its outcome $\textbf{y}$ before inferring the distribution over $\textbf{Z}$. Hence, the following two auxiliary distributions are introduced for the treatment assignment $\textbf{t}$ and the outcome $\textbf{y}$, referring to Equations (\ref{equ_12}), and (\ref{equ_13}).
\begin{equation}
    q(t_i | \textbf{x}_i) = Bern(\sigma(g_{\phi_5}(\textbf{x}_i)))
    \label{equ_12}
\end{equation}
\vspace{-0.4cm}
\begin{equation}
    q(y_i | \textbf{x}_i, t_i) = \sigma(t_{i} g_{\phi_6}(\textbf{x}_i) + (1 - t_{i}) g_{\phi_7}(\textbf{x}_i)),
    \label{equ_13}
\end{equation}
where $g_{\phi_k}(\cdot)$ is a neural network with variational parameters $\phi_k$ for $k=5, 6, 7$. Note that, unlike traditional VAEs simply passing the feature map directly to the latent space, i.e., the upper path of PEM in Figure \ref{fig:figure2}-(a)), the feature map is also sent to the other two paths, i.e., the middle and the lower paths in PEM in Figure \ref{fig:figure2}-(a)), for the posterior estimations of the treatment $t_i$ and the outcome $y_i$.

The introduced auxiliary distributions help us predict $t_i$ and $y_i$ for new samples. To estimate the parameters of these two distributions, $q(t_i | \textbf{x}_i)$ and $q(y_i | \textbf{x}_i, t_i)$, we add an auxiliary objective, referring to Equation (\ref{equ_14}), to the model training objective over $N$ data samples.
\begin{equation}
    \mathcal{L}_{auxiliary} = \nonumber
\end{equation}
\vspace{-0.5cm}
\begin{equation}
    \sum^{N}_{i=1}( \log q(t_i=t_i^{*} | \textbf{x}_i^{*}) + \log q(y_i=y_i^{*} | \textbf{x}_i^{*}, t_i^{*})),
    \label{equ_14}
\end{equation}
where $\textbf{x}_i^*$, $t_i^*$ and $y_i^*$ are observed values in the training set. 

Finally, we have the following overall training objective for the encoder and decoder networks. See Equation (\ref{equ_15}). 
\begin{equation}
    \mathcal{L}_{causal} = \mathcal{L}_{auxiliary} ~ + \nonumber
\end{equation}
\vspace{-0.5cm}
\begin{equation}
    \sum^{N}_{i=1}\mathbb{E}_{q(\textbf{z}_{i}|\textbf{x}_{i}, t_i, y_i)}[\log p(\textbf{x}_i, t_i | \textbf{z}_i) + \log p(y_i|t_i,\textbf{z}_i)  ~ + \nonumber
\end{equation}
\begin{equation}
     \log p(\textbf{z}_i) - \log q(\textbf{z}_i|\textbf{x}_i, t_i, y_i)].
    \label{equ_15}
\end{equation}
See \cite{louizos2017causal} for the detailed derivation and explanation of Equations (\ref{equ_12}) to (\ref{equ_15}), $\log p(\textbf{x}_i, t_i | \textbf{z}_i)$,  $\log p(y_i|t_i,\textbf{z}_i)$, and $\log p(\textbf{z}_i)$.




\noindent\textbf{3.2 Spatial and Channel-wise Attentions}

A dual attention network, based on self-attention, has been proposed to adaptively integrate local features with their global dependencies \cite{fu2019dual,vaswani2017attention}. The dual attention network consists of spatial and channel-wise attention modules. These two modules are capable of modeling the interdependencies in spatial and channel dimensions. To better capture the implicit interactive information, a causal attention mechanism, based on the dual attention network, is introduced to reinforce the probabilistic encoder in Figure \ref{fig:figure2}-(a). In the PEM, the channel-wise attention selectively emphasizes interdependent channel maps by integrating associated features among all channel maps. The spatial attention selectively aggregates the feature at each position by a weighted sum of the features at all positions.

\noindent\textbf{3.3 Video Summary Generation}

In Figure \ref{fig:figure2}-(a), during the inference phase, the input video and input text-based query go through the MFPM for generating feature maps $\textbf{x}$, and the PEM for generating the probabilistic encoding of the feature maps $q(\textbf{z}|\textbf{x}, y, t)$. Then, the generated probabilistic encoding of the feature maps is sent to the probabilistic decoder, referring to Figure \ref{fig:figure2}-(b), to reconstruct $\textbf{x}$, based on $p(\textbf{x}|\textbf{z})$, and generate prediction score labels $y$, based on $p(y|\textbf{z}, t)$, for the input video and query simultaneously. Finally, based on these generated score labels, a set of video frames is selected from the original input video to create the final video summary. Note that the video summary budget is considered as a user-defined hyperparameter in video exploration \cite{huang2020query}.


\section{Experiments}

\begin{figure}[t!]
\begin{center}
\includegraphics[width=1.0\linewidth]{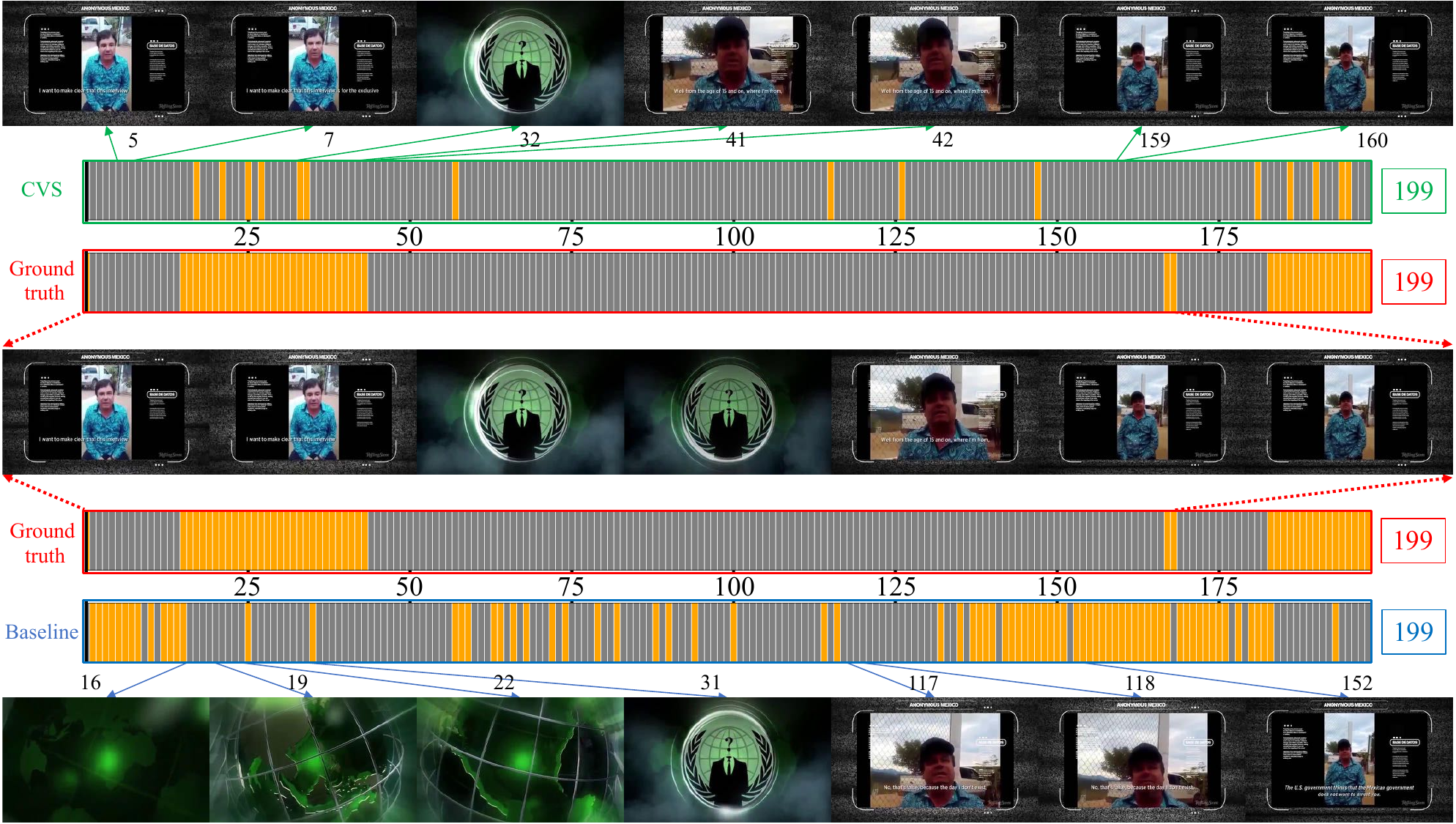}
\end{center}
\vspace{-0.60cm}
   \caption{This figure shows the randomly selected qualitative results of the proposed causal model and the baseline model. According to this figure, we see that the proposed model (in green) prediction result frame pattern is more similar to the ground truth (in red) pattern than the baseline (in blue). Note that in each frame pattern, orange denotes ``not selected frames'' and gray denotes ``selected frames''. Each video here has 199 frames and we also show the corresponding indices of the selected frames in this figure.
   }
\vspace{-0.20cm}
\label{fig:figure3}
\end{figure}

\begin{table*}[t!]
\caption{Ablation study of the introduced attention mechanism and comparison with the state-of-the-art multi-modal video summarization model ``QueryVS'' \cite{huang2020query}. The results are based on the introduced CVSD and the metric of accuracy from \cite{huang2020query}. The attention mechanism is effective and the proposed CVS outperforms the ``QueryVS'' model in the metric of accuracy under six different situations of inputs. Please refer to the \textit{Effectiveness Analysis} section for the definitions of notations in the table.}
\vspace{-0.3cm}
\centering
\scalebox{1.0}{
\begin{tabular}{c|c|cccccc}
\toprule
\multicolumn{2}{c|}{Model} & $V_{s\&p}$ & $V_{blur}$ & $T+V_{s\&p}$ & $T+V_{blur}$ & $T_{k}+V_{s\&p}$ & $T_{k}+V_{blur}$ \\ \hline
\multicolumn{2}{c|}{QueryVS \cite{huang2020query}}    & 0.629  & 0.507  & 0.547  & 0.533  & 0.501    & 0.488  \\ 
\hline
\rowcolor{mygray} \multicolumn{2}{c|}{CVS with attention}       & \textbf{0.687}  & \textbf{0.596}   & \textbf{0.554}  & \textbf{0.535}  & \textbf{0.532}  & \textbf{0.524}  \\ 
\hline
\multicolumn{2}{c|}{CVS without attention}    & 0.676  & 0.423  & 0.442  & 0.530  & 0.514   & 0.500  \\ 
\bottomrule
\end{tabular}}
\label{table:table1}
\vspace{-0.50cm}
\end{table*}

\begin{table}[t!]
	\caption{Comparison with the state-of-the-art model ``QueryVS'' \cite{huang2020query} validated on \cite{huang2020query}'s dataset. The results are based on the metric of accuracy \cite{huang2020query} and $F_{1}$-score \cite{hripcsak2005agreement}. Proposed CVS outperforms the model in \cite{huang2020query} by $+5.4$\% in accuracy and $+4.92$\% in $F_{1}$-score.}
    \vspace{-0.3cm}
	\centering
	\begin{tabular}{c|cc}
		\toprule
		Evaluation Metric & Accuracy & $F_{1}$-score\\
		\hline
		QueryVS \cite{huang2020query} & 0.704 & 0.508\\
		\hline
		\rowcolor{mygray} CVS & \textbf{0.742} & \textbf{0.533}\\
		\bottomrule
	\end{tabular}
	\label{table:table2}
\vspace{-0.5cm}
\end{table}

\noindent\textbf{4.1 Dataset and Evaluation Metrics}

In the experiments, \cite{huang2020query}'s multi-modal video summarization dataset and the introduced causal video summarization dataset (CVSD) with treatment labels are used to verify the proposed method. The introduced CVSD is constructed based on \cite{huang2020query}'s dataset. \cite{huang2020query}'s dataset is composed of $190$ videos with a duration of two to three minutes for each video. Each video in their dataset is retrieved based on a given text-based query. The entire dataset is divided into splits of 60\%/20\%/20\% for training/validation/testing, respectively. Annotations from human experts are necessary for the automatic evaluation of multi-modal video summarization. Hence, the authors of \cite{huang2020query} sample all of the $190$ videos at one frame per second (fps), and Amazon Mechanical Turk (AMT) is used to annotate every frame with its relevance score with respect to the given text-based query. A single ground truth relevance score is created for each query-video pair by merging the corresponding relevance annotations from AMT workers. In \cite{huang2020query}'s dataset, the maximum number of words of a text query is $8$ and the maximum number of frames of video is $199$.

For the evaluation metric, the authors of \cite{huang2020query} propose to use accuracy as the evaluation metric. That is, a predicted relevance score is considered correct when the predicted score is the same as the majority of human experts' provided score. In addition, motivated by \cite{hripcsak2005agreement,song2015tvsum}, in this work $F_{1}$-score is also used to quantify the performance of the proposed model by measuring the agreement between the gold standard and predicted scores provided by the human experts. 

\noindent\textbf{4.2 Experimental Setup}

In this work, \cite{huang2020query}'s dataset and the introduced CVSD have the same preprocessing as follows. Since the video lengths in the CVSD and \cite{huang2020query}'s dataset are varied, the number of frames for each video is different based on fps $=1$. The maximum number of frames of a video is $199$ in the used dataset. Hence, the frame-repeating video preprocessing technique \cite{huang2020query} is adopted to make all the videos have the same length of $199$. ResNet \cite{he2016deep} pre-trained on ImageNet \cite{deng2009imagenet} is used to extract $199$ frame-based features for each video, and the feature used is located in the visual layer one layer below the classification layer. The input frame size of the ResNet is $224$ by $224$ with red, green, and blue channels. Each image channel is normalized by mean $=(0.4280, 0.4106, 0.3589)$ and standard deviation $=(0.2737, 0.2631, 0.2601)$. PyTorch is used for the implementation and to train models with $50$ epochs, $0.01$ learning rate, and Adam optimizer \cite{kingma2014adam}. For the hyperparameters of the Adam, $\beta_{1}=0.9$ and $\beta_{2}=0.999$ are the coefficients used for computing moving averages of gradient and its square. The term added to the denominator to improve numerical stability is $\epsilon=10^{-8}$.

\noindent\textbf{4.3 Effectiveness Analysis}

\noindent\textbf{Effectiveness of Causal Attention Mechanism.}
In the experiments, the CVSD is used to verify the effectiveness of the introduced attention mechanism under the following six different situations of inputs: ``visual input with salt and pepper treatment ($V_{s\&p}$)'', ``visual input with blurring treatment ($V_{blur}$)'', ``clean text-based query input and visual input with salt and pepper treatment ($T+V_{s\&p}$)'', ``clean text-based query input and visual input with blurring treatment ($T+V_{blur}$)'', ``disturbed text-based query input and visual input with salt and pepper treatment ($T_{k}+V_{s\&p}$)'', and ``disturbed text-based query input and visual input with blurring treatment ($T_{k}+V_{blur}$)''. Note that $T_{k}$ denotes randomly removing $k$ words, e.g., $k=2$, from a text-based query input. According to Table \ref{table:table1}, the results show that the introduced attention mechanism results in performance improvement in the causal model. The main reason is that the introduced attention mechanism helps to better capture the implicit interactive information between the visual and textual inputs.

\noindent\textbf{Effectiveness of Causal Effect Modeling.}
The main difference between the proposed causal model and the existing state-of-the-art multi-modal video summarization method, e.g., \cite{huang2020query}, is the causal effect modeling. According to \cite{louizos2017causal}, proper causal effect modeling improves the performance of a machine learning model. In this work, we claim that if the causal effect modeling is effective, the proposed model's performance can be improved.  Based on Table \ref{table:table1} and Table \ref{table:table2}, the results show that the proposed causal model for multi-modal video summarization beats the state-of-the-art \cite{huang2020query}, with the increase of $+5.4$\% in accuracy and $+4.92$\% increase of $F1$-score. The main reason is the proposed model is a causality-based model which properly considers the causal effect modeling. Hence, the claim is well proved by the experimental results. Qualitative results are presented in Figure \ref{fig:figure3}.

\section{Conclusion}  

In this work, a new perspective is introduced to build an end-to-end deep causal model for multi-modal video summarization. The proposed Causal Video Summarizer is based on a probabilistic encoder-decoder architecture. Experimental results show that the proposed causal model is effective and achieves state-of-the-art performance, $+5.4$\% in accuracy and $+4.92$\% increase of $F1$-score. 

\section{Acknowledgments}
This project has received funding from the European Union’s Horizon 2020 research and innovation programme under the Marie Skłodowska-Curie grant agreement No 765140.

\bibliographystyle{IEEEbib}
\bibliography{icme2022template}

\end{document}


\sloppy

\def\x{{\mathbf x}}
\def\L{{\cal L}}

\title{Supplementary}
%
\name{Anonymous ICME submission}
\address{}

\maketitle

\begin{figure*}[t!]
\begin{subfigure}[b]{0.72\textwidth}
\includegraphics[width=1.0\textwidth]{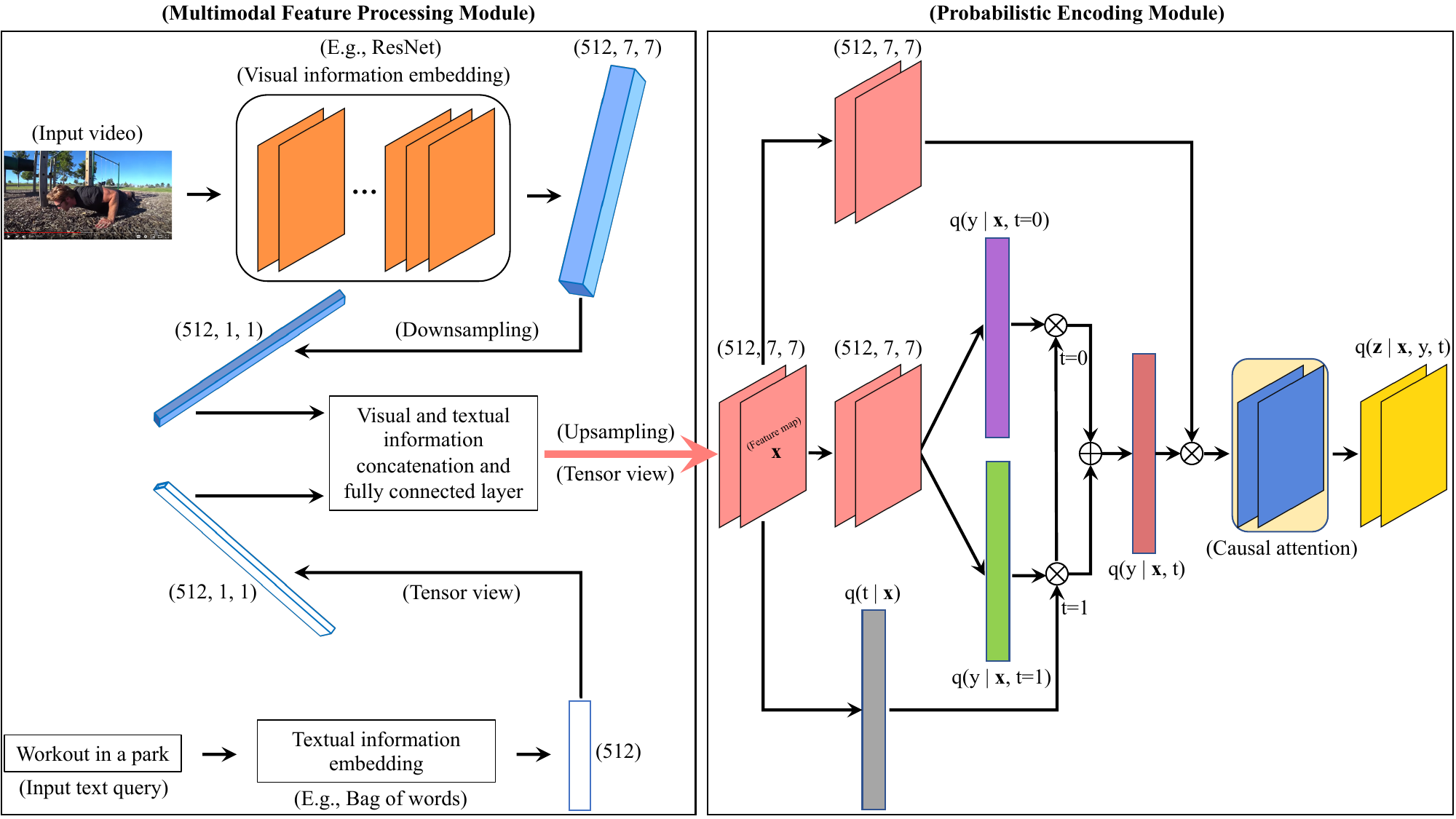}
\caption{Acting as an encoder.}
\end{subfigure}
\begin{subfigure}[b]{0.2585\textwidth}
\includegraphics[width=1.0\textwidth]{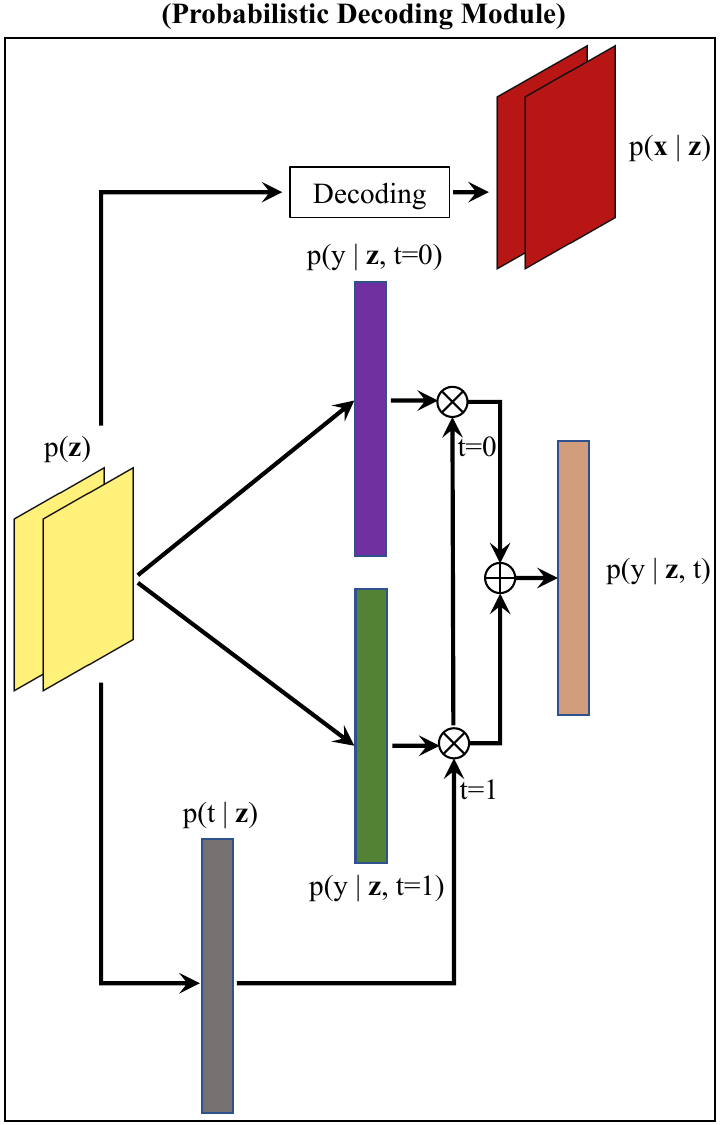}
\caption{Acting as a decoder.}
\end{subfigure}
\vspace{-0.3cm}
   \caption{This figure shows the proposed end-to-end model for multi-modal video summarization, the practical implementation of the causal graphical model in Figure \ref{fig:figure1}. The proposed causal model is composed of multi-modal feature processing module (MFPM), probabilistic encoding module (PEM), and probabilistic decoding module (PDM). In (a), the input video and input text query are processed by MFPM for generating the feature map $\textbf{x}$ which is the input of PEM. Note that, in MFPM, we use 3D views to emphasize the dimension change. Then, PEM generates the corresponding probabilistic encoding $q(\textbf{z}|\textbf{x}, y, t)$ of $\textbf{x}$ with the operations of the introduced attention. Note that, in PEM, pink indicates feature maps from the feature map $\textbf{x}$ and grey denotes the probability distribution of treatment $t$ conditioned on $\textbf{x}$. Green and purple indicate the probability distributions of outcome score labels $y$ conditioned on $(\textbf{x}, t=1)$ and $(\textbf{x}, t=0)$, respectively. Red denotes the probability distribution of outcome score labels $y$ after the fusion of $q(y|\textbf{x}, t=0)$ and $q(y|\textbf{x}, t=1)$ with the operations of summation ($\oplus$) and multiplication ($\otimes$). Light yellow shade indicates the introduced attention and yellow denotes the probabilistic encoding of $\textbf{x}$. In (b), PDM takes $p(\textbf{z})$, estimated by $q(\textbf{z}|\textbf{x}, y, t)$, to approximately reconstruct $\textbf{x}$ based on $p(\textbf{x}|\textbf{z})$ and generate the prediction score labels $y$ based on $p(y|\textbf{z}, t)$ at the same time. Similarly, in PDM, we use different colors to represent different components. Finally, video summaries can be generated based on $y$. Note that, in the context of variational autoencoder \cite{kingma2013auto,louizos2017causal}, (a) is an inference network acting as an encoder; (b) is a model network acting as a decoder.
   }
\vspace{-0.5cm}
\label{fig:figure2}
\end{figure*}

\section{Detailed Derivations of the Proposed CVS}

\subsection{Overview}

In this section, assumptions for modeling multi-modal video summarization are introduced and details of the proposed causal video summarizer are described. The generation of a good video summary is affected by latent factors. In this work, the latent factors are considered as the causal effect. Specifically, the concept of causal effect inference \cite{louizos2017causal}, i.e., a causal graphical model, is used to model the multi-modal video summarization problem, referring to Figure \ref{fig:figure1}. Figure \ref{fig:figure1} contains the four key components of the causal graphical model: ``input data ($\textbf{X}$)'', ``latent confounder ($\textbf{Z}$)'', ``treatment ($\textbf{t}$)'', and ``output result ($\textbf{y}$)''. From the modeling perspective of multi-modal video summarization, $\textbf{X}$ is ``an input video with a text query''. $\textbf{t}$ is ``a visual or textual treatment''. Note that treatment in causal effect modeling is a way to make an input characteristic more salient and help a model learn better \cite{louizos2017causal}. $\textbf{y}$ is ``a relevance score between the input text-based query and video frame or an importance score of a video frame''. $\textbf{Z}$ is ``a variational latent representation'' which the proposed model aims to learn from $\textbf{X}$ for reconstruction \cite{kingma2013auto,louizos2017causal}, referring to Figure \ref{fig:figure2} for the practical implementation. The proposed causal model simultaneously generates outcome score labels $\textbf{y}$ when $\textbf{X}$ is reconstructed, illustrated in Figure \ref{fig:figure2}. Thereafter, video summaries can be created based on the generated outcome score labels. 

\subsection{Assumptions and Causal Video Summarizer}

\begin{figure}[t]
\begin{center}
\includegraphics[width=1.0\linewidth]{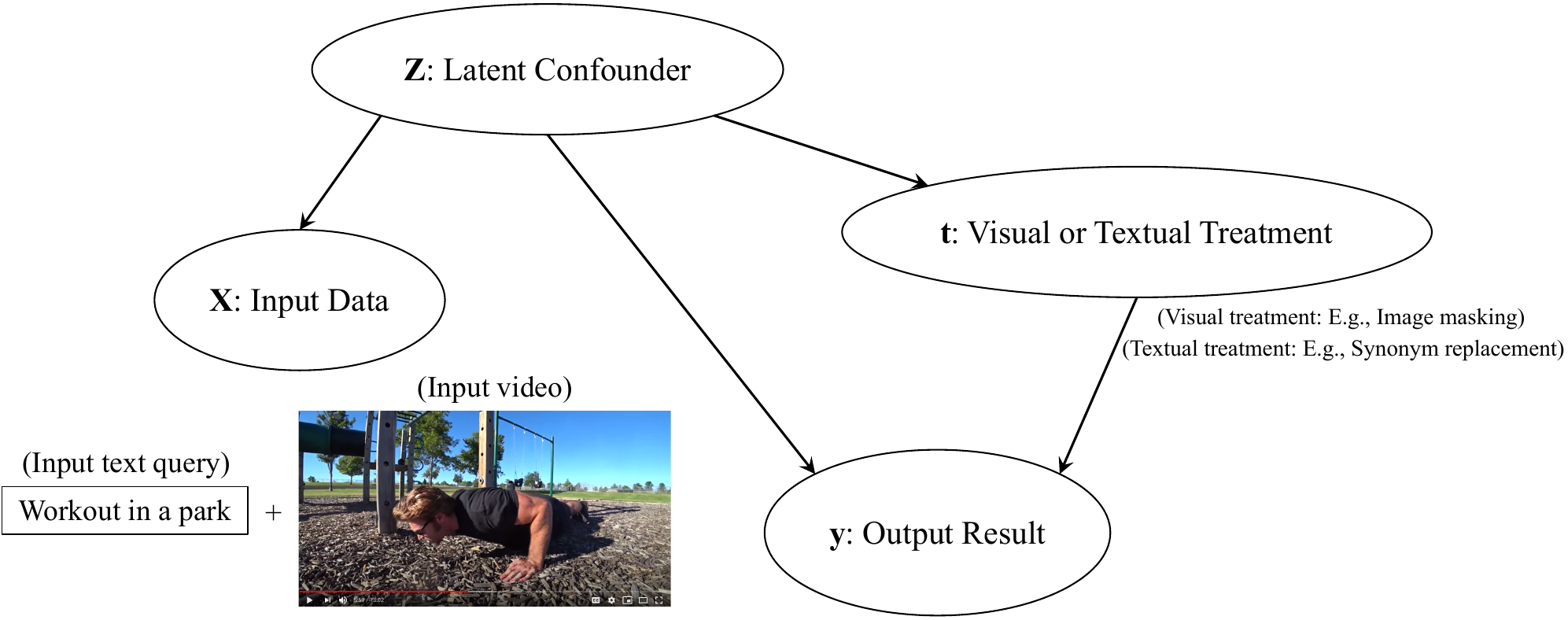}
\end{center}
\vspace{-0.60cm}
  \caption{Explanation of the causal effect modeling concept, i.e., causal graphical model \cite{wickens1972note,cai2012identifying,louizos2017causal}, in multi-modal video summarization. $\textbf{t}$ is a treatment, e.g., visual or textual perturbation. $\textbf{y}$ is an outcome, e.g., an importance score of a video frame or a relevance score between the input text-based query and video. $\textbf{Z}$ is an unobserved confounder, e.g., representativeness or interestingness \cite{gygli2015video,plummer2017enhancing,vasudevan2017query}. $\textbf{X}$ is a noisy view \cite{louizos2017causal} on the hidden confounder $\textbf{Z}$, say the input text query and video.}
\vspace{-0.5cm}
\label{fig:figure1}
\end{figure}

\noindent\textbf{Assumptions.}
Causal effect modeling for real-world multi-modal video summarization is very complicated \cite{louizos2017causal}. Hence, based on the suggestion of \cite{louizos2017causal}, the following two assumptions are imposed when modeling the problem of multi-modal video summarization based on the concept of causal effect inference, i.e., causal graphical model. First, the information of having visual/textual treatment or not, i.e., the treatment factor $\textbf{t}$, is binary. Second, the observations $(\textbf{X}, \textbf{t}, \textbf{y})$ from a deep neural network (DNN) are sufficient to approximately recover the joint distribution $p(\textbf{Z}, \textbf{X}, \textbf{t}, \textbf{y})$ of the latent/unobserved confounders $\textbf{Z}$ and the observed confounders $\textbf{X}$. Specifically, we propose to characterize the causal graphical model in Figure \ref{fig:figure1} as a latent variable model parameterized by a DNN. See Figure \ref{fig:figure2} and the following.   

\noindent\textbf{Causal Video Summarizer (CVS).}
The proposed CVS is mainly composed of a multi-modal feature processing module (MFPM), a probabilistic encoding module (PEM), and a probabilistic decoding module (PDM) as shown in Figure \ref{fig:figure2}. In Figure \ref{fig:figure2}-(a), the purpose of the MFPM is to generate feature maps $\textbf{x}$ with specific tensor dimensions for the multi-modal inputs, i.e., videos and text-based queries. The visual input is processed by a visual information processor, e.g., ResNet34, with a downsampling technique \cite{NEURIPS2019_9015}. The textual input is processed by a textual information processor, e.g., bag of words, with a tensor view operation in PyTorch deep learning framework \cite{NEURIPS2019_9015}. Then, the processed multi-modal information (visual and textual) is concatenated and sent to a fully connected layer with a tensor view operation in PyTorch and an upsampling technique \cite{NEURIPS2019_9015} to create feature maps $\textbf{x}$. In Figure \ref{fig:figure2}-(a), the PEM takes the generated feature maps $\textbf{x}$ as inputs and outputs the corresponding probabilistic encoding $q(\textbf{z}|\textbf{x}, y, t)$ of the feature maps by going through the lower, upper, and middle paths with the attention mechanism applied. The lower path is to generate the probability distribution $q(t|\textbf{x})$ for treatment $t$ based on $\textbf{x}$. The middle path is to generate the probability distribution $q(y|\textbf{x}, t)$ for outcome score labels $y$ based on $\textbf{x}$ and $t$. Then, the feature maps from the upper path and $q(y|\textbf{x}, t)$ with the attention, referring to \textit{Spatial and Channel-wise Attentions} subsection for more details, are used to generate the probabilistic encoding $q(\textbf{z}|\textbf{x}, y, t)$ for latent confounders $\textbf{z}$ based on $\textbf{x}$, $y$, and $t$. There are two main jobs for the PDM, referring to Figure \ref{fig:figure2}-(b). The first job of the PDM is to reconstruct $\textbf{x}$ based on the probability distribution $p(\textbf{x}|\textbf{z})$ which is generated by $p(\textbf{z})$, the priori probability distribution of the latent confounder $\textbf{z}$. The second job of the PDM is to generate prediction score labels $y$, based on the probability distributions $p(y|\textbf{z}, t)$ and $p(t|\textbf{z})$. However, in practice, since we do not have a priori knowledge of the latent confounder $\textbf{z}$, we have to marginalize over it in order to learn the DNN model parameters, referring to Equation (\ref{equ_1}). Then, $q(\textbf{z}|\textbf{x}, y, t)$ is used to approximate $p(\textbf{z})$ based on minimizing the Kullback-Leibler (KL) divergence between two probability distributions \cite{kingma2013auto,rezende2014stochastic}, and then the proposed causal model can be trained after the approximation, referring to the next subsection for details.

\subsection{Objective for Causal Learning}

The authors of \cite{kingma2013auto,louizos2017causal, shalit2017estimating} have shown that variational autoencoder (VAE), which aims to learn a variational latent representation $\textbf{Z}$ from data $\textbf{X}$ for reconstruction, is capable of learning the latent variables when used in a causal graphical model. The authors of \cite{sonderby2016ladder,chen2016variational,tomczak2017vae} have recently expanded the range and type of distributions that can be captured by VAE. Therefore, to effectively infer the complex non-linear relationships between $\textbf{X}$ and $(\textbf{Z}, \textbf{t}, \textbf{y})$ and approximately recover $p(\textbf{Z}, \textbf{X}, \textbf{t}, \textbf{y})$, the proposed multi-modal video summarization model is built on top of VAE.

The proposed method design choices are typical for VAEs: we assume the observations factorize conditioned on the latent variables and use an inference network \cite{kingma2013auto,rezende2014stochastic}, i.e., an encoder, which follows a factorization of the true posterior. For the model network \cite{louizos2017causal, shalit2017estimating}, i.e., a decoder, instead of conditioning on observations we condition on the latent variables $\textbf{z}$, referring to Equations (\ref{equ_1}), (\ref{equ_2}), (\ref{equ_3}), and (\ref{equ_4}). For multi-modal video summarization, $\textbf{x}_i$ corresponds to an input video with a text-based query indexed by $i$, $\textbf{z}_i$ corresponds to the latent confounder, $t_i \in \{0,1\}$ corresponds to the treatment assignment, and $y_i$ corresponds to the outcome. Each of the corresponding factors is described as follows:

\begin{equation}
    p(\textbf{z}_i) = \prod_{z\in \textbf{z}_i} \mathcal{N}(z | \mu=0, \sigma^2=1),
    \label{equ_1}
\end{equation}
where $\mathcal{N}(z | \mu, \sigma^2)$ denotes a Gaussian distribution with a random variable $z$. Note that $z$ is an element of $\textbf{z}_i$. In Equation (\ref{equ_1}), the settings of mean $\mu$ and variance $\sigma^2$ follows the settings in \cite{kingma2013auto}, i.e., $\mu=0$ $\sigma^2=1$.

\begin{equation}
    p(\textbf{x}_i | \textbf{z}_i) = \prod_{x \in \ \textbf{x}_i} p(x | \textbf{z}_i),
    \label{equ_2}
\end{equation}
where $p(x |\textbf{z}_i)$ is an appropriate probability distribution conditioning on $\textbf{z}_i$, with a random variable $x$ which is an element of $\textbf{x}_i$. 

\begin{equation}
    p(t_i | \textbf{z}_i) = Bern(\sigma(f_{\theta_1}(\textbf{z}_i))),
    \label{equ_3}
\end{equation}
where $\sigma(\cdot)$ indicates a logistic function, and $Bern(\cdot)$ denotes a Bernoulli distribution for a discrete outcome.  $f_{\theta_1}(\cdot)$ is a neural network parameterized by the parameter $\theta_1$.

\begin{equation}
    p(y_i | \textbf{z}_i, t_i) = \sigma(t_{i} f_{\theta_2}(\textbf{z}_i) + (1 - t_{i}) f_{\theta_3}(\textbf{z}_i)),
    \label{equ_4}
\end{equation}
where $f_{\theta_2}(\cdot)$ and $f_{\theta_3}(\cdot)$ are neural networks parameterized by the parameters $\theta_2$ and $\theta_3$, respectively. Here $y_i$ is tailored for a categorical classification problem, i.e., relevance score classification in this work, but the formulation can be naturally extended to different tasks. For example, one can simply remove the logistic function $\sigma(\cdot)$ in $p(y_i | \textbf{z}_i, t_i)$ for regression tasks.

Since we do not have a priori knowledge of the hidden confounder $\textbf{z}$, we have to marginalize over it in order to learn the model parameters, $\theta_1$, $\theta_2$, and $\theta_3$. The non-linear neural network functions make inference intractable, so we employ variational inference along with the inference networks, referring to Figure \ref{fig:figure2}-(a). These are neural networks that output the parameters of a fixed form posterior approximation over the latent variables $\textbf{z}$, e.g., a Gaussian distribution, given the observed variables. According to Figure \ref{fig:figure1}, we know that the true posterior over $\textbf{Z}$ depends on $\textbf{X}$, $\textbf{t}$, and $\textbf{y}$. Hence, we employ the following posterior approximation. See Equations (\ref{equ_5}), (\ref{equ_6}), (\ref{equ_7}), (\ref{equ_8}), (\ref{equ_9}), (\ref{equ_10}), and (\ref{equ_11}).

\begin{equation}
    q(\textbf{z}_i | \textbf{x}_i, y_i, t_i) = \prod_{z\in \textbf{z}_i} \mathcal{N}(z | \mu=\bm{\mu}_i, \sigma^2=\bm{\sigma^2}_i)
    \label{equ_5}
\end{equation}

\begin{equation}
    \bm{\mu}_i = t_i \bm{\mu}_{t=1, i} + (1 - t_i) \bm{\mu}_{t=0,i}
    \label{equ_6}
\end{equation}

\begin{equation}
    \bm{\sigma^2}_i = t_i \bm{\sigma^2}_{t=1, i} + (1 - t_i) \bm{\sigma^2}_{t=0, i}
    \label{equ_7}
\end{equation}

\begin{equation}
    \bm{\mu}_{t=0, i} = g_{\phi_1}(g_{\phi_0}(\textbf{x}_i, y_i))
    \label{equ_8}
\end{equation}

\begin{equation}
    \bm{\sigma^2}_{t=0, i} = \sigma(g_{\phi_2}(g_{\phi_0}(\textbf{x}_i, y_i)))
    \label{equ_9}
\end{equation}

\begin{equation}
    \bm{\mu}_{t=1, i} = g_{\phi_3}(g_{\phi_0}(\textbf{x}_i, y_i))
    \label{equ_10}
\end{equation}

\begin{equation}
    \bm{\sigma^2}_{t=1, i} = \sigma(g_{\phi_4}(g_{\phi_0}(\textbf{x}_i, y_i))),
    \label{equ_11}
\end{equation}
where $g_{\phi_k}(\cdot)$ is a neural network with variational parameters $\phi_k$ for $k=0, 1, 2, 3, 4$, and $g_{\phi_0}(\textbf{x}_i, y_i)$ is a shared representation. Specifically, we multiply the feature map with approximated posterior $q(y_i | \textbf{x}_i, t_i)$ without logistic function $\sigma(\cdot)$ to get $g_{\phi_0}(\textbf{x}_i, y_i)$, referring to Figure \ref{fig:figure2}-(a).

Now, it is time to form a single objective for the encoder, in Figure \ref{fig:figure2}-(a), and the decoder, in Figure \ref{fig:figure2}-(b), to learn meaningful causal representations in the latent space and generate video summaries. Based on Figure \ref{fig:figure1}, we know that the true posterior over $\textbf{Z}$ depends on not just $\textbf{X}$, but also on $\textbf{t}$ and $\textbf{y}$. We have to know the treatment assignment $\textbf{t}$ along with its outcome $\textbf{y}$ before inferring the distribution over $\textbf{Z}$. Hence, the following two auxiliary distributions are introduced for the treatment assignment $\textbf{t}$ and the outcome $\textbf{y}$, referring to Equations (\ref{equ_12}), and (\ref{equ_13}).

\begin{equation}
    q(t_i | \textbf{x}_i) = Bern(\sigma(g_{\phi_5}(\textbf{x}_i)))
    \label{equ_12}
\end{equation}

\vspace{-0.1cm}
\begin{equation}
    q(y_i | \textbf{x}_i, t_i) = \sigma(t_{i} g_{\phi_6}(\textbf{x}_i) + (1 - t_{i}) g_{\phi_7}(\textbf{x}_i)),
    \label{equ_13}
\end{equation}
where $g_{\phi_k}(\cdot)$ is a neural network with variational parameters $\phi_k$ for $k=5, 6, 7$. Note that, unlike traditional VAEs simply passing the feature map directly to the latent space, i.e., the upper path of PEM in Figure \ref{fig:figure2}-(a)), the feature map is also sent to the other two paths, i.e., the middle and the lower paths in PEM in Figure \ref{fig:figure2}-(a)), for the posterior estimations of the treatment $t_i$ and the outcome $y_i$.

The introduced auxiliary distributions help us predict $t_i$ and $y_i$ for new samples. To estimate the parameters of these two distributions, $q(t_i | \textbf{x}_i)$ and $q(y_i | \textbf{x}_i, t_i)$, we add an auxiliary objective, referring to Equation (\ref{equ_14}), to the model training objective over $N$ data samples.

\begin{equation}
    \mathcal{L}_{auxiliary} = \nonumber
\end{equation}

\vspace{-0.5cm}
\begin{equation}
    \sum^{N}_{i=1}( \log q(t_i=t_i^{*} | \textbf{x}_i^{*}) + \log q(y_i=y_i^{*} | \textbf{x}_i^{*}, t_i^{*})),
    \label{equ_14}
\end{equation}
where $\textbf{x}_i^*$, $t_i^*$ and $y_i^*$ are observed values in the training set. 

Finally, we have the following overall training objective for the encoder and decoder networks. See Equation (\ref{equ_15}).

\begin{equation}
    \mathcal{L}_{causal} = \mathcal{L}_{auxiliary} ~ + \nonumber
\end{equation}

\vspace{-0.5cm}
\begin{equation}
    \sum^{N}_{i=1}\mathbb{E}_{q(\textbf{z}_{i}|\textbf{x}_{i}, t_i, y_i)}[\log p(\textbf{x}_i, t_i | \textbf{z}_i) + \log p(y_i|t_i,\textbf{z}_i)  ~ + \nonumber
\end{equation}

\begin{equation}
     \log p(\textbf{z}_i) - \log q(\textbf{z}_i|\textbf{x}_i, t_i, y_i)].
    \label{equ_15}
\end{equation}

Note that the evident lower bound (ELBO), \cite{kingma2013auto,rezende2014stochastic,louizos2017causal}, of the proposed causal graphical model to be optimized is as shown in Equation (\ref{equ_16}).

\begin{equation}
    ELBO =  \mathcal{L}_{causal} - \mathcal{L}_{auxiliary}.
    \label{equ_16}
\end{equation}

\begin{figure}[t!]
\begin{center}
\includegraphics[width=1.0\linewidth]{dataset_example_s&p.pdf}
\end{center}
\vspace{-0.50cm}
   \caption{A randomly selected example with visual treatment, i.e. salt and pepper disturbance in this case, from the introduced CVSD. The first row denotes an original video. The second row denotes the video with visual treatment.}
\vspace{-0.30cm}
\label{fig:figure4}
\end{figure}

\begin{figure}[t!]
\begin{center}
\includegraphics[width=1.0\linewidth]{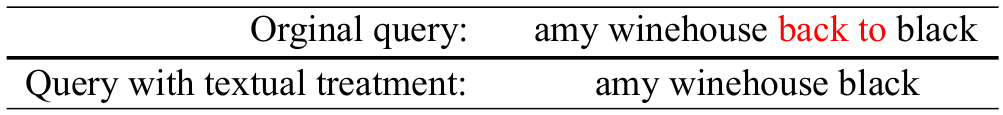}
\end{center}
\vspace{-0.50cm}
   \caption{A randomly selected example with textual treatment, i.e., randomly removing $k=2$ words in this case, from the introduced CVSD. The first row denotes an original query. The second row denotes the query with textual treatment. The difference between the first and second rows is ``back to'', which is randomly removed.}
\label{fig:figure5}
\end{figure}

\subsection{Causal Video Summarization Dataset (CVSD)}

Recently, the authors of \cite{huang2020query} have proposed a multi-modal video summarization dataset, including videos and corresponding meaningful text-based queries, with 
frame-based score labels.
The length of each video is with a duration of two to three minutes and the score labels are annotated by Amazon Mechanical Turk workers. The dataset is composed of 114 training, 38 validation, and 38 testing videos. Based on \cite{huang2020query}'s dataset with the same training, validation, and testing sets, the CVSD is made through the following steps. First, 
we randomly select 50\% of the \textit{(video, query)} data pairs from the original training, validation, and testing sets. 
Secondly, for each selected video, we assign 0, i.e., without treatment, or 1, i.e, with treatment, treatment labels to 30\% of the video frames and the corresponding queries. Finally, we use the same method mentioned in \cite{huang2020query,sharghi2018improving} to assure each video has the same length.

According to \cite{louizos2017causal}, in causal effect modeling, adding a treatment to an input is a way to make the input characteristic more salient and help a model learn better. In this work, treatments for visual and textual inputs are acquired based on the following: When we observe people's writing behaviors, we notice that some patterns recur often, such as synonym replacement, accidentally missing some words in a sentence, and so on. Motivated by this, we pick one common behaviour, e.g., accidentally missing some words in a sentence, and write a textual treatment function to simulate it. Similarly, we know that when people make videos in their daily life, some visual disturbances are common, such as salt and pepper noise, image masking, blurring, and so on. We also randomly pick up some of them, e.g., blur and salt and pepper noise, and make a visual treatment function as a simulation. Based on the visual and textual simulation functions, we introduced the CVSD with visual and textual treatments. For the introduced CVSD dataset example, please refer to Figure \ref{fig:figure4} and Figure \ref{fig:figure5}. Note that in the real world, there are various common disturbances. The aforementioned randomly selected visual and textual treatments are just a few of them. The other common treatments also can be used in the proposed method. Now, we can train the causal model on the CVSD.

\bibliographystyle{IEEEbib}
\bibliography{icme2022template}